\documentclass{article}
\usepackage{spconf,amsmath,graphicx}

\usepackage{epsfig}
\usepackage{graphicx}
\usepackage{subfigure}
\usepackage{epstopdf}
\usepackage{amsmath}
\usepackage{amssymb}
\usepackage{array}
\usepackage{xcolor}
\usepackage{booktabs}
\usepackage{multirow}
\usepackage{algorithm}
\usepackage{algorithmic}

\usepackage[utf8x]{inputenc} 

\title{MULTI-STAGE RESIDUAL HIDING FOR IMAGE-INTO-AUDIO STEGANOGRAPHY}
\name{Wenxue Cui$^{\star}$ \quad Shaohui Liu$^{\star}$ \quad Feng Jiang$^{\star}$ \quad Yongliang Liu$^{\dagger}$ \quad Debin Zhao$^{\star}$}

\address{
 $^{\star}$ Harbin Institute of Technology, Harbin, China \quad $^{\star}$ Peng Cheng Laboratory, Shenzhen China \\
 $^{\dagger}$ Alibaba Group, Hangzhou, China }
\begin{document}
%
\maketitle
\begin{abstract}
The widespread application of audio communication technologies
has speeded up audio data flowing across the Internet,
which made it a popular carrier for covert communication.
In this paper, we present a cross-modal steganography
method for hiding image content into audio carriers while
preserving the perceptual fidelity of the cover audio. In our
framework, two multi-stage networks are designed: the first
network encodes the decreasing multilevel residual errors
inside different audio subsequences with the corresponding
stage sub-networks, while the second network decodes the
residual errors from the modified carrier with the corresponding
stage sub-networks to produce the final revealed results.
The multi-stage design of proposed framework not only
make the controlling of payload capacity more flexible, but
also make hiding easier because of the gradual sparse characteristic
of residual errors. Qualitative experiments suggest
that modifications to the carrier are unnoticeable by human
listeners and that the decoded images are highly intelligible.
\end{abstract}
\begin{keywords}
Audio steganography, residual hiding, multi-stage network, convolutional neural networks
\end{keywords}
\section{Introduction}
\label{sec:intro}

Sometimes a file carries more information than it perceptively conveys. To a casual observer, a file may appear normal, but knowledgeable recipients can extract more information from it. Recently, to protect confidential data, steganography
has been researched extensively, which is a technique of concealing secret messages in digital carriers to facilitate covert
communication through exploiting the redundancy of human perceptions. The secrecy characteristics of steganography are attractive for diverse applications such as copyright certification~\cite{s1} and covert communication~\cite{s2}.

Faithful hiding heavily depends on its embedding method. To achieve perfect hiding performance, a wide variety of steganography setting and methods have been proposed. For instance, Least Significant Bit (LSB) substitution methods~\cite{s3,s4,s5} have been extremely popular for steganography due to their simplicity. Then some advanced methods emerged such as HUGO~\cite{s6}, WOW~\cite{s7} and S-UNIWARD~\cite{s8}, which embedded the messages in complex textures by minimizing a well-crafted distortion function and achieved superior performance. However, the aforementioned methods usually use domain knowledge to identify features for hiding secret message into cover carriers, which usually leads to small payload capacities and apparent distortion altering.

In recent years, deep neural networks (DNN) have recently been applied to steganography, with a strong focus on images. Instead of specifying domain knowledge explicitly, DNN-based methods have been explored to learn the signal characteristics implicitly. The earliest application of deep
learning to steganography was based on Generative Adversarial Networks (GAN)~\cite{s9,s10,s11,s12,s13}. The introduction
of the adversarial training not only resists more kinds of attacks but also achieves better visual performance. In order to enhance the payload capacities of steganography, several steganography algorithms~\cite{s1,s14,s15} that embed an image into another image are proposed. Furthermore, with the popularization
of diverse audio applications, researchers begin to
pay more attention to speech steganography. In~\cite{s16}, two neural networks are optimized jointly to embed message into the cover audio and extract message from the modified carrier. In~\cite{s17}, a GAN-based audio-to-audio framework is proposed, in which an encoder network and a decoder network is designed
for information embedding and extraction in the frequency domain by using short-time fourier transform.

Compared with the hand-crafted embedding methods, the aforementioned deep network-based steganography methods have achieved superior performance. However, these frameworks still exist the following weaknesses: (1) Most of these methods seem to be powerless for the steganography between different data modalities. (2) Hiding the secret message directly is difficult because of its diversity of knowledge and therefore usually leads to noticeable artifacts.

\begin{figure*}
\centering
\includegraphics[width=\textwidth]{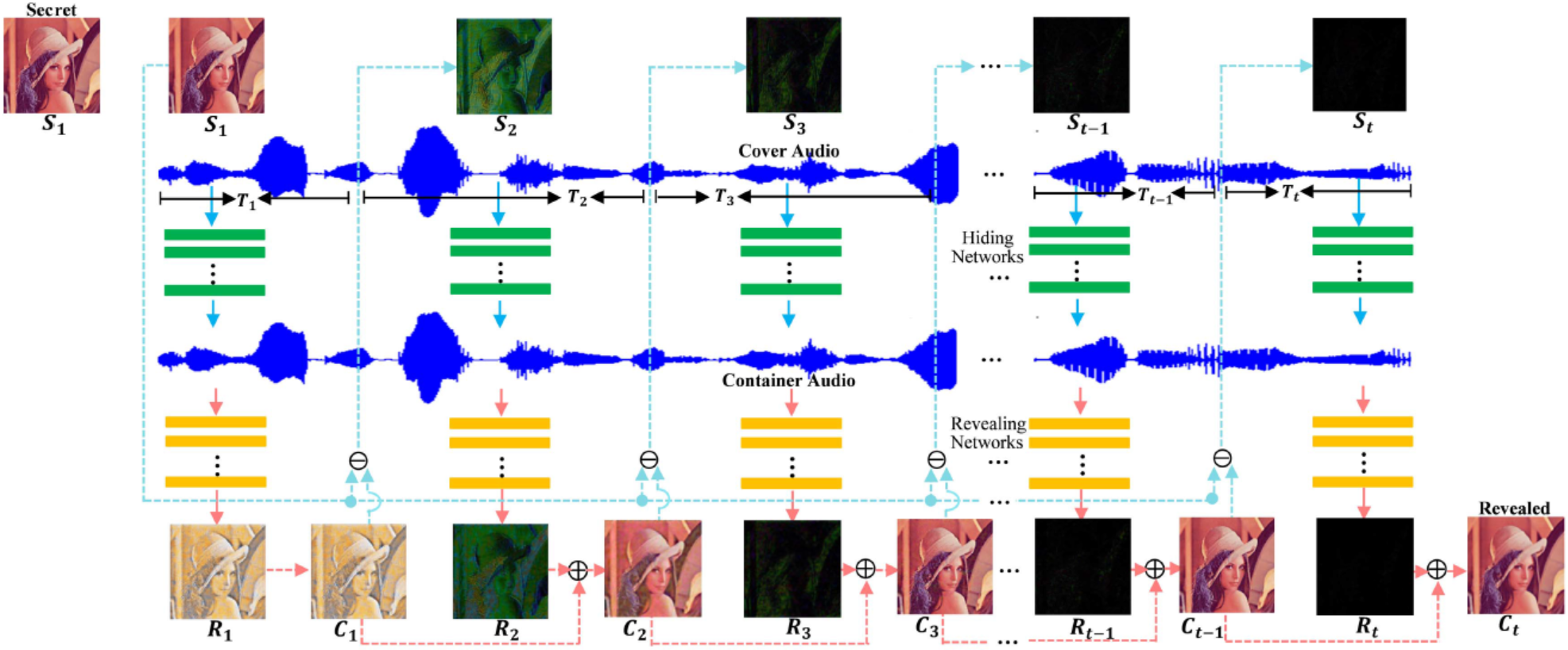}
\vspace{-0.36in}\caption{Diagram of our proposed steganography framework.
}
\vspace{-0.21in}
\label{Fig:framework}
\end{figure*}

To overcome the shortcomings of the aforementioned
methods, we propose a Deep neural network based Image-To-Audio Steganography (DITAS) framework as shown in Fig.1, in which two multi-stage networks are designed: hiding network and revealing network. Specifically, the hiding network encodes the decreasing residual errors inside different audio subsequences with the corresponding stage
sub-networks, and the revealing network decodes the residual errors from the modified carrier with the corresponding stage sub-networks to produce the final revealed results. The proposed framework hides the secret message in an progressive
fashion, which makes hiding process more easier. Experimental
results on benchmark datasets demonstrate that the
proposed method is capable of achieving superior performance
against other algorithms.

The main contributions are summarized as follows:

\textbf{1)} We propose a novel image-to-audio steganography
framework based on deep learning, which achieves superior
hiding capabilities against other methods.

\textbf{2)} By hiding the residual errors of multiple levels, the proposed
method not only can control the payload capacity more
flexibly, but also make the hiding process more easier.

\textbf{3)} Our framework embed the residual errors into different
audio subsequences, which implies that even if part of the
carrier is lost, the secret image can be restored to some extent.

\section{Proposed Method}
\label{sec:format}

\subsection{Hiding Network}

Given a secret image $S_{0}$ with size $w$$\times$$h$ and a cover audio
$A_{l}$, where $w$, $h$ are the width and height of secret image
and $l$ indicates the dimension of cover audio. We embed
multilevel residual errors of secret image into cover audio
progressively by using a multi-stage network. Specifically, $t$
non-overlapping audio subsequences are first selected from
the cover audio sequence, which expressed mathematically
as $\{$$T_{1}$, $T_{2}$,..., $T_{t}$$\}$ and the dimension of each subsequence is
$w$$*$$h$. The proposed framework consists of $t$ stages to embed
residual errors of the secret image into these $t$ subsequences
correspondingly. More concretely, for the $i$-$th$ stage, we hide
the residual error between the original secret image and the
revealed results from the previous $i$-$1$ stages revealing subnetworks.
The hiding process can be expressed as
\begin{equation}
\tilde{T}_{i}=\mathcal{H}(S_{i}, T_{i}; \theta_{H_{i}})
\end{equation}
where $\mathcal{H}$ indicates the operation of hiding network and $\theta_{H_{i}}$
is the parameter of $i$-$th$ stage hiding sub-network. $S_{i}$ is the
$i$-$th$ level residual error of secret image to be hidden and $\tilde{T_{i}}$
is the hidden result (Container) that preserves the perceptual
fidelity of the cover audio $T_{i}$. The residual error $S_{i}$ can be
expressed as
\vskip -0.2in
\begin{equation}
S_{i}=S_{0}-C_{i-1}, \quad C_{i-1}=\sum_{j=0}^{i-1}R_{j}
\end{equation}
where $S_{0}$ is the original secret image, $R_{i}$ ($i$ $>$ 0) indicates
the revealed result of $i$-$th$ stage revealing sub-network, which
will be demonstrated in the next subsection and $R_{0}$ is a zero
tensor with the same size of $S_{0}$. $C_{i}$ indicates the total sum
of revealed results of previous $i$ stages revealing sub-networks.

\begin{figure}[b]
\centering
\vspace{-0.2in}
\includegraphics[width=3.3in]{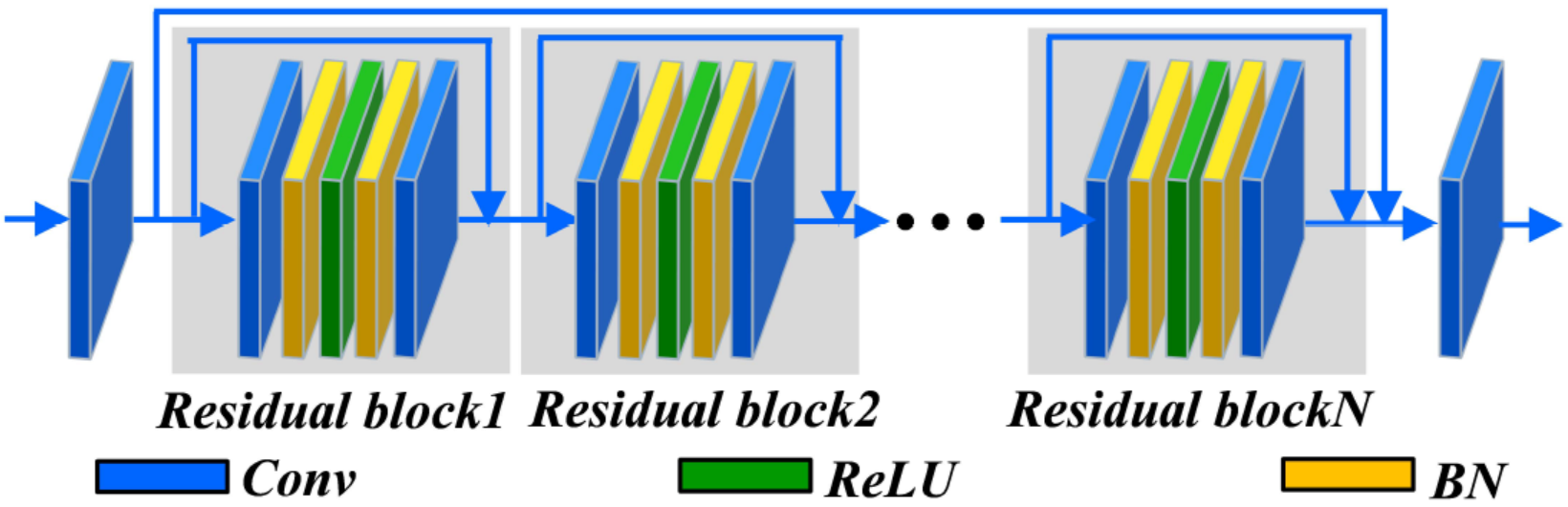}
\vspace{-0.18in}\caption{The architecture of proposed residual block.}
\vspace{-0.06in}
\label{Fig:framework}
\end{figure}

\begin{figure*}
\centering
\includegraphics[width=\textwidth]{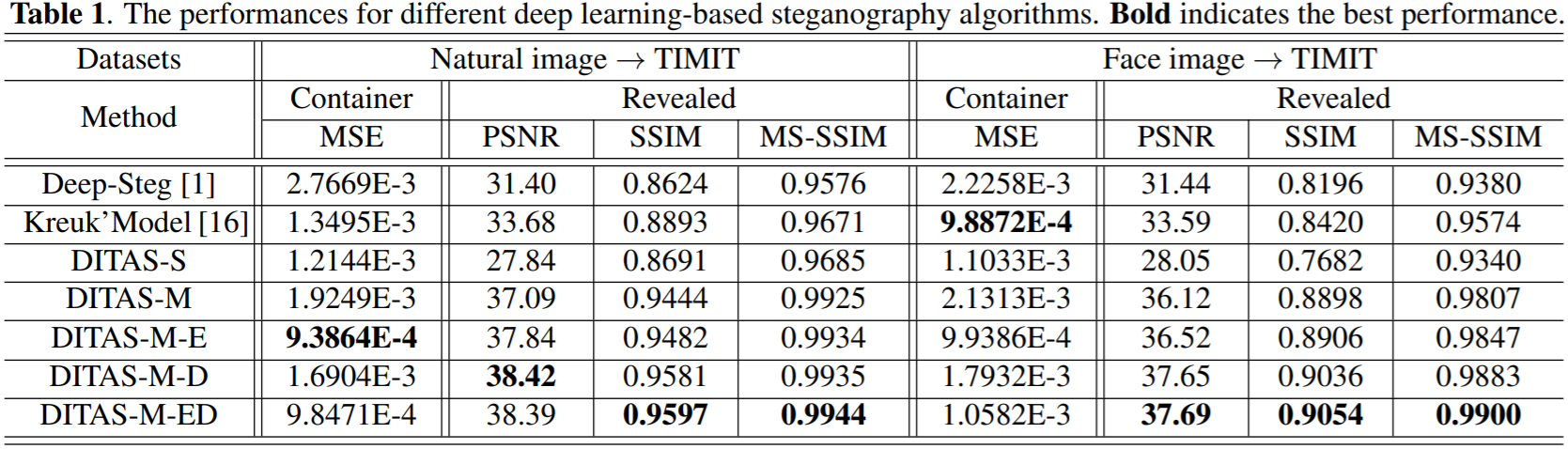}
\vspace{-0.3in}
\label{Fig:framework}
\end{figure*}

\subsection{Revealing Network}

After hiding stage, the hidden audio subsequence $\tilde{T_{i}}$ is produced.
Subsequently, the goal of revealing network is to extract
the secret image from the hidden audio sequence in the
revealing stage. Specifically, given the hidden audio subsequence $\tilde{T_{i}}$, the revealing process can be expressed as
\begin{equation}
R_{i}=\mathcal{R}(\tilde{T_{i}};\theta_{R_{i}})
\end{equation}
where $\mathcal{R}$ indicates the operation of revealing network and $\theta_{R_{i}}$
is the parameter of its $i$-$th$ stage revealing sub-network. $R_{i}$
is the revealed residual result from $\tilde{T}_{i}$. After get the residuals
$R_{i}$, we add them together to produce the final revealed
result, namely $C_{i}=\sum_{j=0}^{i}R_{j}$. It is worth emphasizing that
the proposed framework embeds multi-level residuals in different
audio subsequences and therefore the revealed result is
extracted progressively from the container sequence. In other
words, the subsequences are independent between each other
and even if some subsequences are lost, the secret image can
also be revealed to some extent, which improves the robustness
of the proposed method.

\begin{figure}[h]
\centering
\vspace{-0.1in}
\includegraphics[width=3.3in]{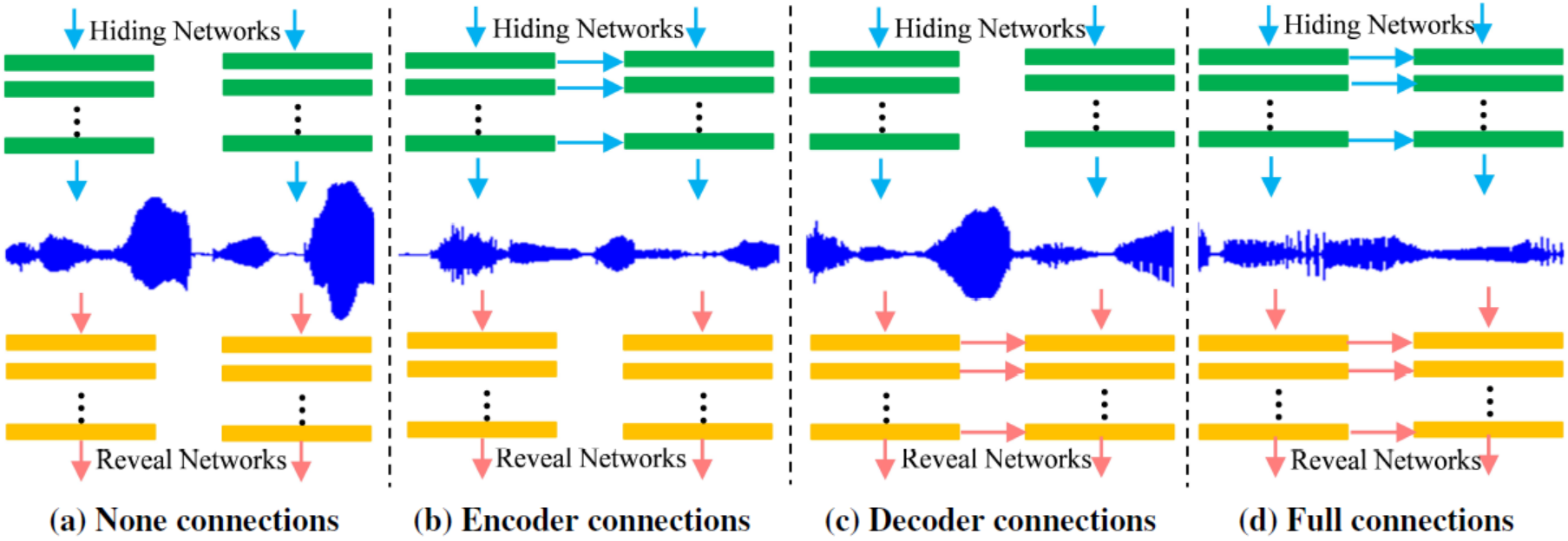}
\vspace{-0.06in}\caption{ The structural details of four experimental variants.}
\vspace{-0.16in}
\label{Fig:framework}
\end{figure}

\vspace{-0.1in}
\subsection{Loss Function}

In the proposed framework, we have two main missions: one
is secret image hiding, the other is secret image extraction.
Therefore, there are two loss items to constraint the container
and the extracted entities, respectively. Besides, the proposed
framework consists of $t$ stages and therefore the complete loss
function can be expressed as:
\vskip -0.16in
\begin{equation}
\mathcal{L}(\theta_{H_{i}},\theta_{R_{i}})=\sum_{i=1}^{t}\mathcal{L}_{H_{i}}(\theta_{H_{i}})+\lambda_{i}\mathcal{L}_{R_{i}}(\theta_{R_{i}})
\end{equation}
where $\mathcal{L}_{H_{i}}$
is hiding loss of $i$-$th$ stage hiding sub-network
for information concealing and $\mathcal{L}_{R_{i}}$
is revealing loss of $i$-$th$
stage revealing sub-network for information extraction. $\theta_{H_{i}}$
and $\theta_{R_{i}}$
are the parameters of them. $\lambda_{i}$
is the regularization
parameter to control the tradeoff between them. Specifically,
the hiding loss is defined as
\begin{equation}
\mathcal{L}_{H_{i}}(\theta_{H_{i}})=\frac{1}{N}\sum_{i=1}^{N}\| \mathcal{H}(S_{i}, T_{i};\theta_{H_{i}})-T_{i}\|_{2}^{2}
\end{equation}
On the other hand, to ensure the precision of extracted information, another extraction loss is defined
\begin{equation}
\mathcal{L}_{R_{i}}(\theta_{R_{i}})=\frac{1}{N}\sum_{i=1}^{N}\| \mathcal{R}(\tilde{T_{i}};\theta_{R_{i}})-S_{i}\|_{2}^{2}
\end{equation}
where $\tilde{T_{i}}=\mathcal{H}(S_{i}, T_{i};\theta_{H_{i}})$ indicates the container audio sequences.

\begin{figure}[b]
\centering
\vspace{-0.2in}
\includegraphics[width=3.3in]{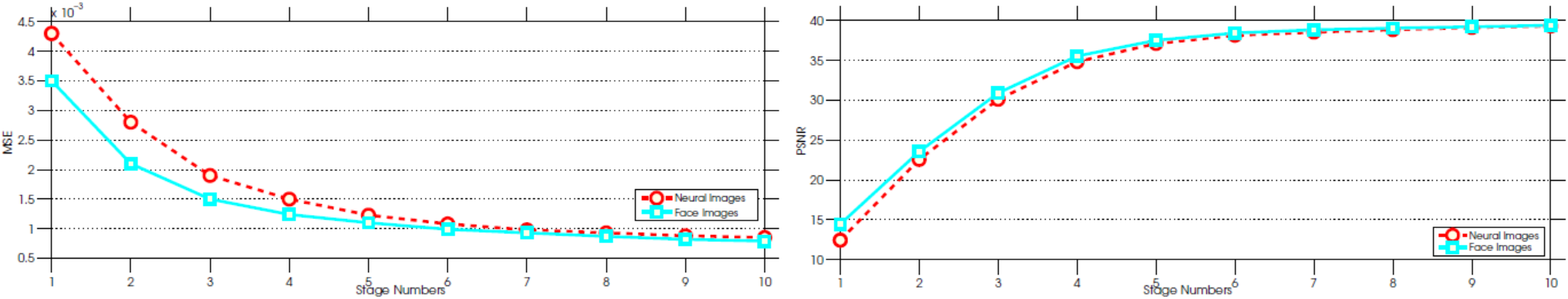}
\vspace{-0.18in}\caption{The relationship between the number of stages and
results based on audio dataset TIMIT. ( Left is the MSE of
container audio and right is PSNR of revealed image).}
\vspace{-0.06in}
\label{Fig:framework}
\end{figure}

\vspace{-0.1in}

\section{EXPERIMENTAL RESULTS AND ANALYSIS}
\label{sec:pagestyle}

\subsection{ Implementation and training details}
In order to embed the secret image into the cover audio by
convolutional operations, preprocessing is needed for the one-dimensional audio subsequence $T_{i}$
. Specifically, two methods
of audio data pre-processing are conducted: 1) Raw audio data is reshaped directly to a new tensor. 2) Short-Time Fourier
Transform (STFT) is applied to transform audio from the audio domain to the frequency domain.

After the pre-processing, a tensor with size of $w$$\times$$h$ is
produced for $T_{i}$
, which can be concatenated with the secret
entity $S_{i}$
together conveniently as the input of $i$-$th$ stage hiding sub-network. In the proposed hiding and revealing networks, the proposed residual block is utilized as shown in
Fig.2. Specifically, for each stage hiding sub-network, each
convolutional layer consists of 64 kernels with size of 3$\times$3
except for the last layer, in which single kernel is included to
ensure dimensional consistency between the output and the
cover audio tensor. Similarly, for the revealing network, the
last convolutional layer has 3 kernels to extract the final revealed residuals and 64 kernels are included for the rest layers. We train our model with the python toolbox Pytorch~\cite{s18}
on a Titan X GPU. Adaptive moment estimation (Adam)~\cite{s19} is utilized to optimize all network parameters. Furthermore,
we set $\lambda_{i}$= 0.8. In order to choose the optimal number of
stages ($t$), we train a model with 10 stages and Fig.4 shows
the relationship between the performance and the stage numbers, from which we can see that with the increase of stages,
the performance growth is gradually slow down and we set
$t$ = 5 in our model. It is worth emphasizing that the proposed
framework can be trained in an end-to-end fashion.

For training, we use the training set of dataset VOC2012~\cite{s20} as the secret image and use the LJ Speech dataset~\cite{s21}
as the cover audio. For testing, we choose 2 kinds of images as the secret image: natural image (including Set14~\cite{s22},
LIVE1~\cite{s23}, Val of VOC2012~\cite{s20} and Val of ImageNet~\cite{s14})
and face image (including CelebA dataset~\cite{s12}) , which are
widely used in the literatures. Besides, we use the TIMIT audio dataset~\cite{s24} as the cover audio. For training details, the
patch size is set as 64$\times$64, which is cropped from training
dataset randomly and we set batch size as 16. The learning
rate is initialized to $1e$-4 for all layers and decreased by a
factor of 3 for every 20 epochs. We train the entire network
for 200 epochs.

\begin{figure}[t]
\centering
\includegraphics[width=3.3in]{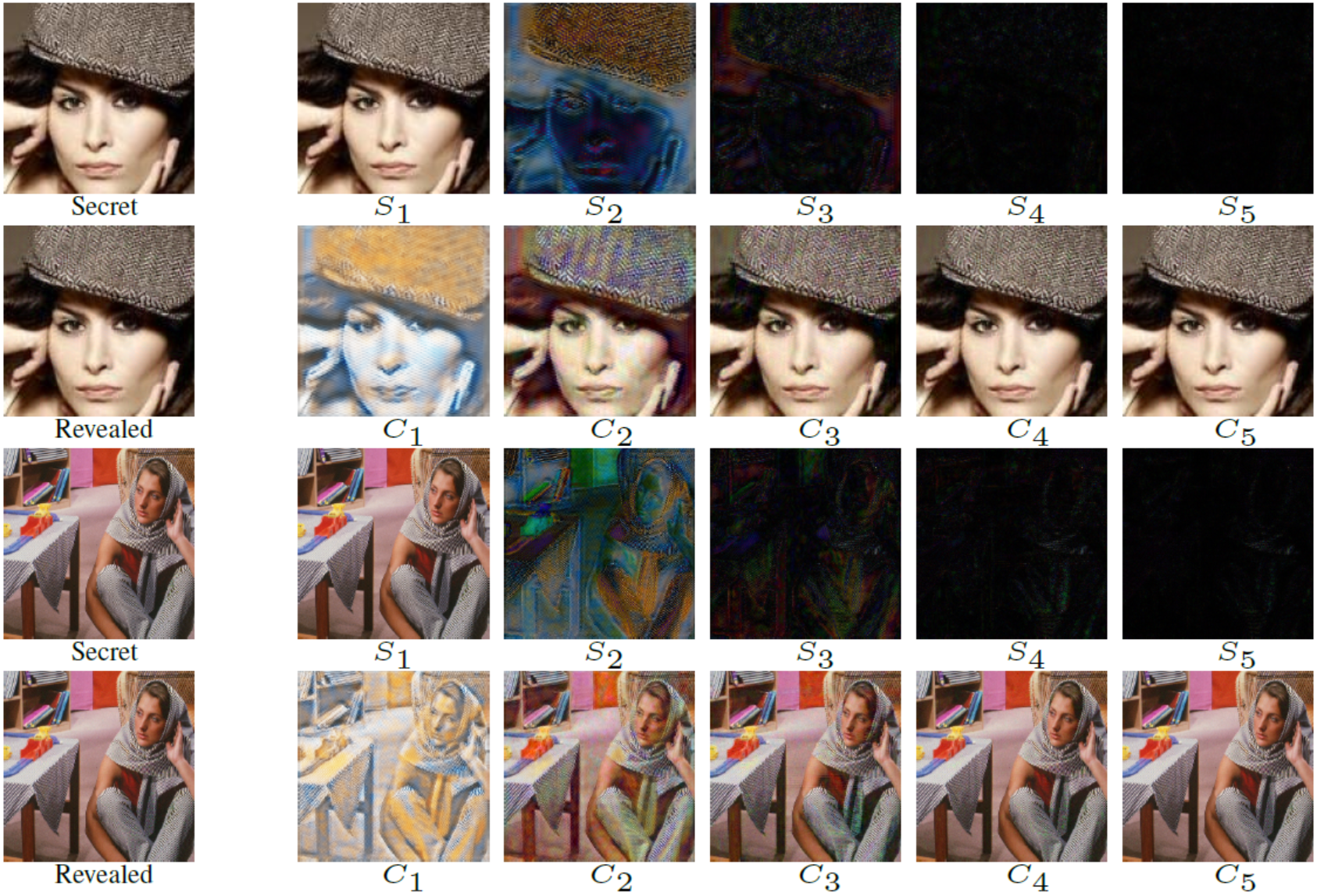}
\vspace{-0.18in}\caption{The intermediate visual outputs of our framework.}
\vspace{-0.16in}
\label{Fig:framework}
\end{figure}

\vspace{-0.1in}

\subsection{ Comparison with other methods}

Following the deep steganography structure proposed by
Baluja et al.~\cite{s1}, we can build a network model for hiding
an image in audio. Specifically, given a cover audio, $t$ non-overlapping sub-sequences with size of $w$$*$$h$ are first selected.
For each subsequence, we reshape it as $w$$\times$$h$ and concatenate
them together, after which we get a tensor with size $t$$\times$$w$$\times$$h$.
Then, we embed the secret image into this tensor. Besides,
refer to the previous work~\cite{s16}, we can also obtain the audio
tensor by using STFT, and then embed the secret image into
the frequency domain of the cover audio.

Considering the proposed framework, we try two preprocessing methods mentioned in subsection 3.1 for the
cover audio transformation. We find that steganography
in frequency domain is slightly better than that in raw audio
data domain. While the execution of STFT will increase
the computational complexity. Therefore, we reshape the
raw audio data into tensor directly in this work. In the proposed framework, we train each stage subnetwork without
any information transfer between them (shown in Fig.3(a))
and name it as ``DITAS-M''. Besides, to verify the efficiency
of information transfer between stages, four experimental
variants are designed: 1) We build the connections between
successive stages of hiding subnetworks to propagate information (shown in Fig.3(b)) and we name it as ``DITAS-M-E''.
2) The connections between successive stages of revealing
subnetwork are builded (shown in Fig.3(c)) and we name it
as ``DITAS-M-D''. 3) Full connections are set up between
neighbouring stages (shown in Fig.3(d)) and we name it as
``DITAS-M-ED''. 4) Shared parameters are utilized for each
stage and we name it as ``DITAS-S''.

For experimental verification, we use Mean Square Error
(MSE) to measure the distortion of hidden audio and utilize
three evaluation metrics (PSNR, SSIM, MS-SSIM) to measure the quality of revealed image. Table.1 shows the objective experimental results, from which we can see that the
proposed method can achieve superior performance against
other methods and obtains more than 3dB gains in terms of
PSNR. Fig.5 shows the intermediate outputs of our model,
from which we can see that $S_{i}$
is more and more sparse and
the vision of $C_{i}$
is better and better. Fig.6 shows the comparisons of subjective quality between the proposed method and
others, from which we can see that the proposed method is
capable of achieving better visual performance.

\begin{figure}[h]
\centering
\vspace{-0.14in}
\includegraphics[width=3.3in]{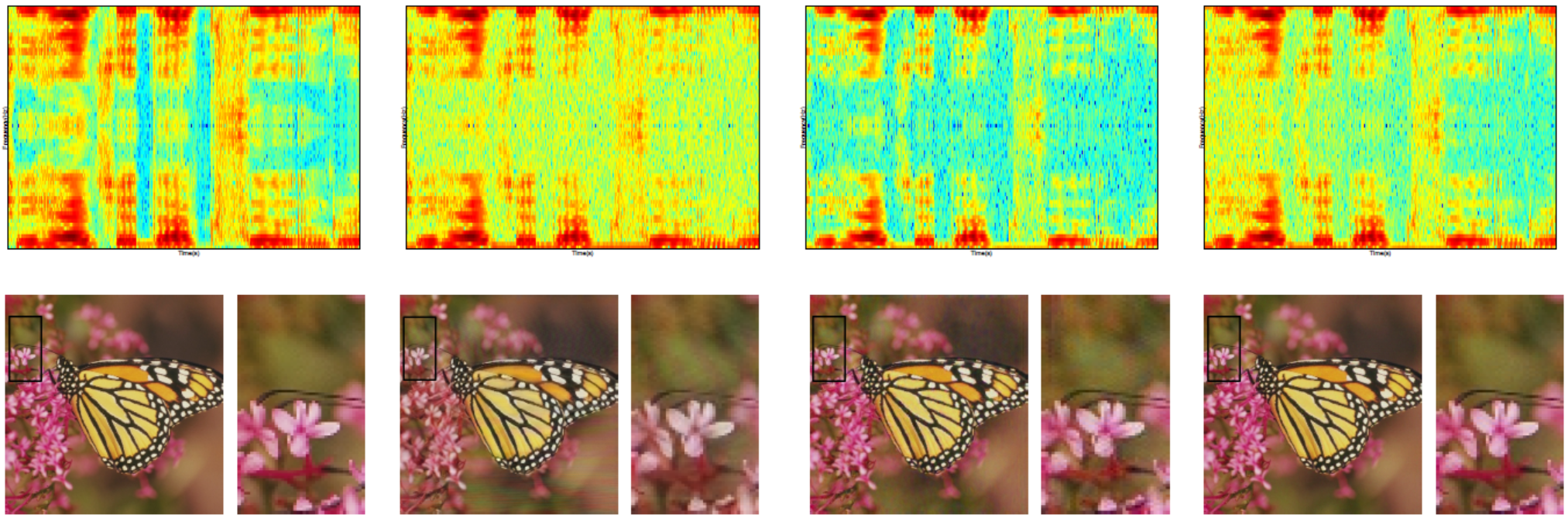}
\vskip -0.58in \tiny{Original \ \quad\quad\quad\quad\quad\quad\quad Deep-Steg \ \quad\quad\quad\quad\quad\quad\quad Kreuk's Model \quad\quad\quad\quad\quad\quad\quad \ Ours}
\vspace{0.40in}\caption{The visual comparisons of different steganography
methods. (Top is the visual comparisons of container in the
frequency domain and bottom is the revealed perceptions.)}
\vspace{-0.19in}
\label{Fig:framework}
\end{figure}

\vspace{-0.1in}

\section{CONCLUSION}
\label{sec:typestyle}

In this paper, we propose a novel cross-modal image-to-audio
steganography framework based on deep learning. Instead
of hiding the secret image directly, the proposed method embeds the residual errors of secret image into the cover audio
progressively by a multi-stage fashion. In the hiding process
of the proposed method, residual errors become more sparse
with the increase of stages, which not only make the controlling of payload capacity more flexible, but also make hiding
easier because of the sparsity characteristic of residual errors.


\vspace{-0.1in}
\section{ACKNOWLEGEMENT}
\label{sec:refs}
This work was supported by Alibaba Group through Alibaba Innovative Research (AIR) Program and partly funded by National Key Research and
Development Program of China via grant 2018YFC0806802
and 2018YFC0832105.

\bibliographystyle{IEEEbib}
\bibliography{strings,refs}

\end{document}